\documentclass[letterpaper]{article} 
\usepackage{aaai2026}
\usepackage{times}  
\usepackage{helvet}  
\usepackage{courier}  
\usepackage[hyphens]{url}  
\usepackage{graphicx} 
\usepackage{booktabs}
\usepackage{tabularx}
\usepackage{multirow} 
\usepackage{threeparttable}
\usepackage{makecell}
\usepackage{amssymb}
\usepackage{booktabs}
\usepackage{tabularx}
\usepackage{colortbl}
\usepackage{multirow} 
\usepackage{xcolor}
\usepackage{courier} 
\usepackage{nicematrix}
\usepackage{makecell}
\usepackage{natbib}  
\usepackage{caption} 
\usepackage{algorithm}
\usepackage{algorithmic}
\usepackage{amsmath}
\usepackage{amssymb}
\usepackage{amsfonts}
\usepackage{cases}
\usepackage{newfloat}
\usepackage{listings}
\DeclareCaptionStyle{ruled}{labelfont=normalfont,labelsep=colon,strut=off} 
\floatstyle{ruled}
\newfloat{listing}{tb}{lst}{}
\floatname{listing}{Listing}
\title{ME$^3$-BEV: Mamba-Enhanced Deep Reinforcement Learning\\for End-to-End Autonomous Driving with BEV-Perception}
\author{
    Siyi Lu\textsuperscript{\rm 1,\rm 2},
    Run Liu\textsuperscript{\rm 3},
    Dongsheng Yang\textsuperscript{\rm 4},
    Lei He\textsuperscript{\rm 1,\rm 2}\thanks{Corresponding author: helei2023@tsinghua.edu.cn}
}
\affiliations{
    \textsuperscript{\rm 1}School of Vehicle and Mobility, Tsinghua University, Beijing 100084, China.\\

    \textsuperscript{\rm 2}State Key Laboratory of Intelligent Green Vehicle and Mobility, Tsinghua
University, Beijing 100084, China.\\
    \textsuperscript{\rm 3}Research Center of Ubiquitous Sensor Networks, University of Chinese Academy of Sciences, Beijing 100049, China.\\
    \textsuperscript{\rm 4}BYD Company Limited, Shenzhen, 518118, China.
}

\usepackage{bibentry}

\begin{document}

\maketitle

\begin{abstract}
Autonomous driving systems face significant challenges in perceiving complex environments and making real-time decisions. Traditional modular approaches, while offering interpretability, suffer from error propagation and coordination issues, whereas end-to-end learning systems can simplify the design but face computational bottlenecks. This paper presents a novel approach to autonomous driving using deep reinforcement learning (DRL) that integrates bird's-eye view (BEV) perception for enhanced real-time decision-making. We introduce the \texttt{Mamba-BEV} model, an efficient spatio-temporal feature extraction network that combines BEV-based perception with the Mamba framework for temporal feature modeling. This integration allows the system to encode vehicle surroundings and road features in a unified coordinate system and accurately model long-range dependencies. Building on this, we propose the \texttt{ME$^3$-BEV} framework, which utilizes the \texttt{Mamba-BEV} model as a feature input for end-to-end DRL, achieving superior performance in dynamic urban driving scenarios. We further enhance the interpretability of the model by visualizing high-dimensional features through semantic segmentation, providing insight into the learned representations. Extensive experiments on the CARLA simulator demonstrate that \texttt{ME$^3$-BEV} outperforms existing models across multiple metrics, including collision rate and trajectory accuracy, offering a promising solution for real-time autonomous driving.

\end{abstract}

\section{Introduction}

\begin{figure}
    \centering
    \includegraphics[width=1\linewidth]{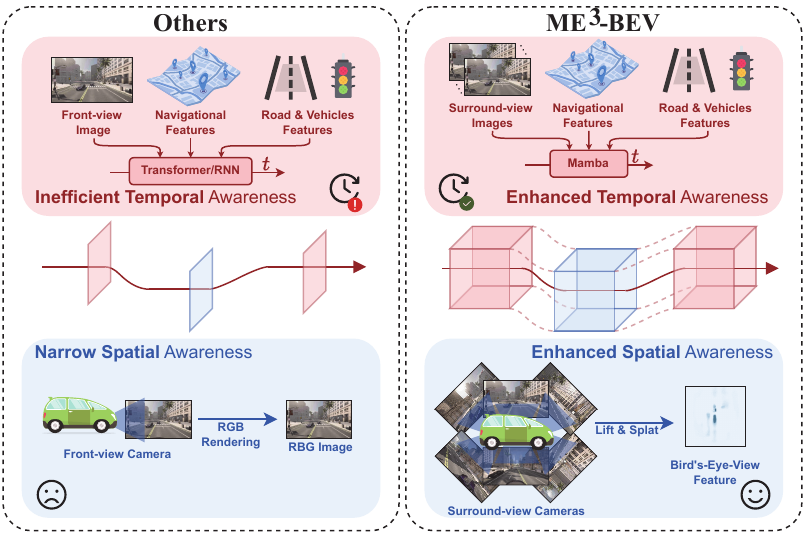}
    \caption{\texttt{ME$^3$-BEV} is designed to enhance spatial perception and leverage temporal information, aiming to reduce collisions and enable accurate trajectory following.}
    \label{fig:motivation}
\end{figure}

Autonomous driving has emerged as a transformative technology that has the potential to revolutionize transportation by improving safety, efficiency, and accessibility~\cite{narisetty2024revolutionizing,zhao2024autonomous}. To achieve fully autonomous driving in complex real-world environments, a key challenge lies in designing control systems that can perceive the environment, understand the driving context, and make safe and efficient decisions in real time~\cite{10614862,chib2023recent}.

Existing approaches to autonomous driving are generally categorized into two paradigms: modular pipelines and end-to-end learning. The modular approach decomposes the driving task into separate perception, planning, and control modules, allowing each to be optimized independently. Although this design promotes interpretability and modularity, it suffers from accumulated errors between modules and complex data coordination. In contrast, end-to-end learning aims to map sensor input directly to driving actions using deep neural networks, which can simplify the system architecture and reduce error propagation. Recent works have shown the promise of deep reinforcement learning (DRL) in learning robust end-to-end driving policies in complex urban scenarios~\cite{10316275,su2024compressing}.

However, existing end-to-end autonomous driving methods face several key limitations. First, they struggle to effectively and efficiently process long-term dependencies in sequential sensor inputs in real time. Current methods typically employ recurrent neural networks (RNNs)~\cite{fei2025data}, which are limited in capturing long-range dependencies, or Transformers~\cite{vaswani2017attention}, which incur significant inference latency and hinder real-time deployment. Second, very few studies incorporate bird’s-eye view (BEV) representations into deep reinforcement learning for end-to-end autonomous driving policies while also meeting real-time inference requirements~\cite{li2024bevformer}. 

Moreover, DRL-based driving strategies often lack interpretability and have limited perception of the surrounding environment, increasing the risk of collisions and compromising safety~\cite{zhao2024survey}.Additionally, current DRL approaches struggle to fully exploit long-horizon sequential input features, limiting their ability to accurately follow planned trajectories~\cite{xiao2024decision}. Accurate trajectory execution relies heavily on understanding the intentions of other traffic participants, which in turn requires modeling temporal behavior over extended time spans.

To address these challenges, we propose a framework that enhances BEV perception to improve spatial understanding and integrates a Mamba-based temporal module to better capture sequential dependencies. As shown in Figure~\ref{fig:motivation}, our design aims to improve both the safety and the overall quality of autonomous driving control. The main contributions of our work are as follows.

\begin{itemize}
    \item We propose \texttt{Mamba-BEV} model, an efficient spatio-temporal feature extraction network that integrates BEV perception and the Mamba framework. The network encodes the vehicle's surround camera images into BEV features and aligns road and vehicle features via a coordinate system. These features are then fed into a temporal feature extraction network composed of Mamba to obtain the final spatio-temporal representation.
    \item We propose an end-to-end deep reinforcement learning framework for autonomous driving, named \texttt{ME$^3$-BEV}, which integrates \texttt{Mamba-BEV} for feature input. To enhance the interpretability of \texttt{ME$^3$-BEV}, we also introduce a method that visualizes the high-dimensional features fed into \texttt{ME$^3$-BEV} as semantic outputs.
    \item We conducted extensive evaluations in seven different maps in the CARLA simulator, comparing the performance of \texttt{ME$^3$-BEV} with the baseline in terms of driving score, collision rate, and other key metrics. \texttt{ME$^3$-BEV} achieved superior performance across multiple metrics.
\end{itemize}

\begin{figure*}[t]
    \centering
    \includegraphics[width=0.93\linewidth]{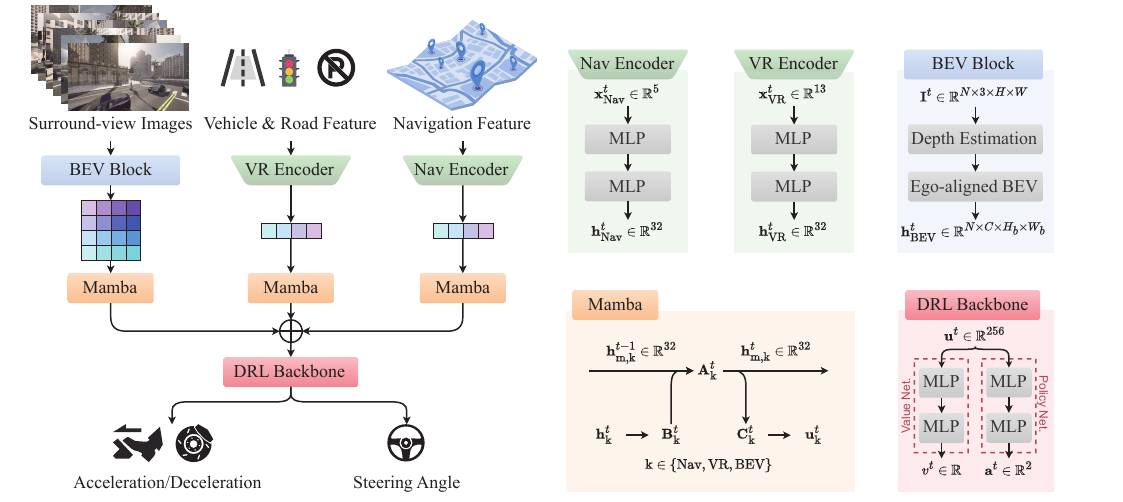}
    \caption{Overview of the proposed \texttt{ME$^3$-BEV} architecture. Surround-view camera images, road features, and navigation information are used as inputs. A BEV encoder extracts spatial features from the surround-view images, while the MAMBA module captures temporal dependencies. The deep reinforcement learning framework outputs final vehicle control commands.}
    \label{fig:arch}
\end{figure*}

\section{Related Work}
Existing work in autonomous driving can be divided into two types: Modular autonomous driving system and end-to-end autonomous driving system.

\noindent
\textbf{Modular Autonomous Driving System.} 
In the modular autonomous driving system, the autonomous driving system is decomposed into several distinct modules, each responsible for specific functions such as perception, planning, decision-making, and control~\cite{zhao2025survey,tang2023survey}. The perception module equips the vehicle with the ability to sense its surroundings, enhancing capabilities in object detection~\cite{chen2017multi}, semantic segmentation~\cite{nesti2022evaluating}, lane detection~\cite{hou2019learning}, and pedestrian recognition~\cite{hsu2020ratio}. The planning module is responsible for path planning, generating trajectory points for the vehicle's low-level controller~\cite{ji2023tripfield, li2024integrated}. The decision-making and control module executes low-level control commands to ensure vehicle stability and safety~\cite{kong2015kinematic}. While each module focuses on a specific task, thereby reducing the overall complexity of the autonomous driving system, coordinating data transmission and management between modules remains a challenge, necessitating effective monitoring and integration. In contrast, the end-to-end autonomous driving system can effectively address these coordination challenges.

\noindent
\textbf{End-to-End Autonomous Driving System.} 
In end-to-end autonomous driving system, driving policies are made directly from raw sensor inputs, enabling high computational efficiency. DRL has emerged as a highly effective approach for learning end-to-end autonomous driving policies, delivering impressive performance across various driving scenarios~\cite{kiran2021deep,grigorescu2020survey,zhu2021survey}. There are many works  have explored the application of DRL in this domain \cite{aradi2020survey}. For instance, Chen et al. \cite{chen2019model} introduced a framework tailored to support model-free DRL in complex urban driving environments. In another work, \cite{9196730}, reinforcement learning was employed within a simulation platform to develop a driving system capable of controlling a full-scale real-world vehicle. Their policy leverages RGB images and their corresponding semantic segmentation from a monocular camera as input. Additionally, \cite{8911507} presented an integrated decision-making framework that combines classical planning techniques with learning-based methods. By uniting Monte Carlo Tree Search with DRL, their approach effectively tackles challenges such as environmental complexity, sensor noise, and nuanced interactions with dynamic road agents. The perception modules in existing works are typically based on GRU, LSTM, or Transformer architectures, which either struggle to capture long-term dependencies or suffer from slow inference speed, making them unsuitable for end-to-end autonomous driving systems with strict real-time requirements~\cite{han2023collaborative,sun2025oe}. In this work, we design an efficient end-to-end autonomous driving model that meets the high real-time demands of control decision-making in autonomous driving scenarios.

\section{Methodology}

\subsection{Problem Formulation} 
Our work aims at addressing end-to-end autonomous driving problem that ego-vehicle reaches a given target location while avoiding collisions with other vehicles and the control decisions are made in real-time. This problem can be formulated as a Partially Observable Markov Decision Process (POMDP)~\cite{Hafner2020Dream, xiang2021recent}. It can be represented by a tuple $\left \langle \mathcal{A},\mathcal{S},\mathcal{R},\mathcal{P},\mathcal{O},\mathcal{Z},\gamma \right \rangle $, where $\mathcal{A}$ represents the action space, $\mathcal{S}$ is the state space, $R$ denotes the reward function, $\mathcal{P}$ represents the state transition function, $\mathcal{O}$ denotes the observation space, $\mathcal{Z}$ is the observation function, and $\gamma$ is the discount factor. In the context of end-to-end autonomous driving, the state transition function $\mathcal{P}$ and the observation function $\mathcal{Z}$ are often not available in closed-form, rendering this a model-free POMDP problem.
\begin{itemize}
    \item \textit{State Space} $\mathcal{S}$: We use the CARLA driving simulator~\cite{dosovitskiy2017carla} as the environment for our agent. The state space $\mathcal{S}$ is determined by CARLA's internal simulation, but it is not directly observable by the agent. At each time step $t$, the environment's state is denoted by $s_t$.
    \item \textit{Observation Space} $\mathcal{O}$: Following the setting in~\cite{e2e-cla,anzalone2021reinforced}, the agent's observation at time step $t$ is defined as $o_t=\bigcup_{m=1}^M{\left\{ \left< I,R,V,N \right> _m \right\}}$, where each tuple $\langle I, R, V, N \rangle_m$ corresponds to the observation from the $m$-th entity. Specifically, $I$ consists of six RGB images (front and rear views), each with a shape of $3\times128\times352$, resulting in a combined tensor of shape $6\times3\times128\times352$. $R \in \mathbb{R}^9$ represents road-related features, $V \in \mathbb{R}^4$ encodes vehicle-related information, and $N \in \mathbb{R}^5$ contains navigation-related features. Here, we use $\mathbf{x}_{\mathrm{VR}}^t$ and $\mathbf{x}_{\mathrm{Nav}}^t$ represent the vehicle-road features and navigation features at time step $t$, respectively. $\mathbf{I}^t$ represents the inputted images. 
    \item \textit{Action Space} $\mathcal{A}$: The agent's actions consist of a continuous steering angle and a longitudinal control value (representing either acceleration or braking). All action values lie within the range $[-1, 1]$.
    \item \textit{Reward Function} $\mathcal{R}$: The reward function is defined as:
    \begin{equation}
        \mathcal{R}\left( s_t,a_t \right) =r_t=\begin{cases}
    	-k_c,\;\;\;\;\;\;\;\;\mathrm{if}\;\mathrm{collision};\\
    	v_m-v_c,\;\;\mathrm{if}\;v_c-v_m>0;\\
    	\frac{4v_c\cdot v_s}{\left\| p_c-p_w \right\| _{2}^{2}},\;\mathrm{otherwise},\\
        \end{cases}
    \end{equation}
    where $v_m$ is the maximum allowed speed (i.e., the speed limit), $v_c$ is the current speed of the vehicle, $v_s$ is a similarity measure between the vehicle's current heading and the direction toward the next waypoint $w$, $k_c$ is a penalty constant for collisions, $p_c$ and $p_w$ denote the positions of the vehicle and the next waypoint, respectively.
\end{itemize}

\subsection{DRL-Based End-to-End Autonomous Driving} 
DRL has emerged as a powerful framework for addressing POMDPs, which naturally model the uncertainties inherent in autonomous driving. In this work, we adopt the Proximal Policy Optimization (PPO) algorithm~\cite{schulman2017proximal} as our core DRL method, owing to its stability and efficiency in continuous control settings. Our framework follows an Actor–Critic architecture (see Figure~\ref{fig:arch}), with policy inputs composed of multimodal observations, including navigation data, vehicle and road state, and multi-view RGB images from a surround-view camera system. Each modality is processed through a dedicated encoder. Specifically, the navigation input $\mathbf{x}_{\mathrm{Nav}}^t$ is encoded via:
\begin{equation}
\mathbf{h}_{\mathrm{Nav}}^t=\mathrm{NavEncoder}(\mathbf{x}_{\mathrm{Nav}}^t),
\end{equation}
while the combined vehicle-road state vector $\mathbf{x}_{\mathrm{VR}}^t$ is embedded as:
\begin{equation}
\mathbf{h}_{\mathrm{VR}}^t=\mathrm{VREncoder}(\mathbf{x}_{\mathrm{VR}}^t).
\end{equation}
The image observation $\mathbf{I}^t$, composed of six surround-view RGB frames, is processed using a BEV-based convolutional network named BEV Block:
\begin{equation}
\mathbf{h}_{\mathrm{BEV}}^t=\mathrm{BEVBlock}(\mathbf{I}^t).
\end{equation}
To capture temporal dependencies across frames, each modality's encoded feature is further processed using the Mamba sequence model:
\begin{equation}
\mathbf{u}_{\mathrm{Nav}}^t=\mathrm{Mamba}(\mathbf{h}_{\mathrm{Nav}}^t,\mathbf{h}_{\mathrm{m,Nav}}^{t-1}),
\end{equation}
\begin{equation}
\mathbf{u}_{\mathrm{VR}}^t=\mathrm{Mamba}(\mathbf{h}_{\mathrm{VR}}^t,\mathbf{h}_{\mathrm{m,VR}}^{t-1}),
\end{equation}
\begin{equation}
\mathbf{u}_{\mathrm{BEV}}^t=\mathrm{Mamba}(\mathbf{h}_{\mathrm{BEV}}^t,\mathbf{h}_{\mathrm{m,BEV}}^{t-1}),
\end{equation}
The fused spatio-temporal representation is computed by summing the modality-specific embeddings:
\begin{equation}
    \mathbf{u}^t=\mathbf{u}_{\mathrm{Nav}}^t+\mathbf{u}_{\mathrm{VR}}^t+\mathbf{u}_{\mathrm{BEV}}^t.
\end{equation}
This integrated feature vector is then forwarded to both the Actor and Critic networks for action generation and value estimation, respectively. By combining BEV Block with Mamba-based temporal modeling, our system is able to construct a consistent and structured understanding of its environment, which significantly improves decision-making in dynamic driving scenarios. Details of the BEV Block module are provided in the next section.

\subsection{Spatial-Semantic Aggregator (SSA)}

Traditional image feature extraction methods typically operate within the original camera coordinate frame, without accounting for transformations into a shared perception space. However, in autonomous driving, most perception modules—such as lidar and map-based components—rely on features represented in the BEV coordinate system. This mismatch between coordinate frames leads to inconsistencies during sensor fusion. To address this, we adopt a Lift-Splat–based BEV encoder~\cite{philion2020lift}, which transforms multi-view RGB images into a unified BEV representation. The process involves two main steps: \textit{Lift} and \textit{Splat}.

In \textit{Lift} step, given an input image $I \in \mathbb{R}^{3 \times H \times W}$, the network first predicts a depth distribution $\alpha_{h,w} \in \Delta_{|D|-1}$ for each pixel location $(h, w)$ over a set of discrete depth bins $D = \{d_1, d_2, \dots, d_n\}$. These depth probabilities are used to lift the 2D image features into 3D space:
\begin{equation}
    I(h,w,d)=\sum_{d\in D}\alpha_{h,w}(d)\cdot f_{h,w}(d),
\end{equation}
where $f_{h,w}(d)$ is the feature vector at depth $d$, and $\alpha_{h,w}(d)$ is the likelihood of the pixel corresponding to that depth. This process results in a frustum-shaped 3D point cloud per pixel, providing a depth-aware encoding of the scene geometry.

In \textit{Splat} step, these 3D features are then projected onto a BEV grid using the camera's intrinsic and extrinsic parameters. Each 3D point is transformed into the ego-vehicle coordinate frame via a mapping function $\mathrm{EGO}_{Ts}$, and assigned to a voxel using sum pooling:
\begin{equation}
    b_i=\sum_i{\mathrm{EGO}_{Ts}\left( u_i \right)},
\end{equation}
where $u_i$ is the context vector of a 3D point. This step aggregates features from all cameras into a spatially consistent BEV representation. By lifting 2D features into 3D and aligning them in a unified coordinate space, this approach enables effective fusion of multi-view image data. The resulting BEV representation preserves geometric consistency, is robust to calibration errors, and supports translation equivariance and permutation invariance—key properties for downstream tasks such as planning and obstacle avoidance.

\subsection{Temporal-Aware Fusion Module (TAFM)}
The TAFM is a type of State Space Model (SSM) that provides a framework for capturing and modeling the temporal dynamics of a system~\cite{smith2023simplified, bradbury2016quasi, lei2017simple}. There are two main equations in TAFM: the state equation and the observation equation. The state equation can be expressed as
\begin{equation}
    \mathbf{h}_{m,k}^t=\mathbf{A}_k^t \mathbf{h}_{m,k}^{t-1}+\mathbf{B}_k^t \mathbf{h}_{k}^t,
\end{equation}
where $\mathbf{h}_{m,k}^t$ is the state vector of $k\in\{\mathrm{Nav, VR, BEV}\}$ at time step $t$, $\mathbf{A}_k^t$ is the state transition matrix, $\mathbf{B}_k^t$ is the control-input matrix, and $\mathbf{h}_{k}^t$ is the input feature vector of $k$. $\mathbf{A}_k^t$ and $\mathbf{B}_k^t$ are trainable. The observation equation models the system's dynamics by describing the relationship between the current state, the input features, and the subsequent state. The observation equation can be expressed as
\begin{equation}
    \mathbf{u}_k^t=\mathbf{C}_k^t \mathbf{h}_{m,k}^t+\mathbf{v}_t,
\end{equation}
where $\mathbf{u}_k^t$ is the embedding vector of the input feature vector of $k$, $\mathbf{C}_k^t$ is the observation matrix related $k$, and $\mathbf{v}_t$ is a additive noise. The observation equation describes how the system's underlying state is manifested in the observed measurements.

The TAFM integrates an input-dependent mechanism~\cite{gu2024mambalineartimesequencemodeling} that dynamically adjusts model parameters based on the current inputs. Additionally, it incorporates a selective mechanism aimed at reducing computational complexity, allowing the model to focus on relevant information while ignoring irrelevant details. This selective focus enhances the model's ability to capture complex patterns, particularly in autonomous driving scenarios. The the state transition matrix $\mathbf{A}_k^t$, the control-input matrix $\mathbf{B}_k^t$, and the observation matrix $\mathbf{C}_k^t$ are mapping of $\mathbf{h}_{k}^t$, i.e.,
\begin{equation}
    \mathbf{A}_k^t = f_A(\mathbf{h}_{k}^t),\quad\mathbf{B}_k^t = f_B(\mathbf{h}_{k}^t),\quad\mathbf{C}_k^t = f_C(\mathbf{h}_{k}^t),
\end{equation}
where $f_A$, $f_B$, and $f_C$ are learnable functions that transform the input $\mathbf{h}_{k}^t$ into the corresponding matrices. These functions can be implemented using neural networks, which allow the model to capture intricate, input-dependent transformations.

\subsection{Semantic Segmentation of Latent Features}

The quality of the input feature representations in DRL significantly affects downstream decision-making performance. However, the relationship between latent features and final task performance often remains implicit and underexplored. To bridge this gap, we introduce a decoding mechanism that maps the latent BEV features into semantically interpretable outputs using a semantic segmentation task. Semantic segmentation plays a fundamental role in autonomous driving, offering pixel-level scene understanding that enables accurate recognition of road structure, dynamic agents, and obstacles. In our framework, we leverage this task to assess and visualize the representations learned by the BEV Feature Extraction Network. Specifically, we design a decoder to transform the high-dimensional latent feature $\mathbf{u}_{\text{BEV}}^t$ into a semantic map $\mathbf{S}$.
\begin{align}
\nonumber
    \mathbf{S} &= \mathrm{Decoder}(\mathbf{u}_{\text{BEV}}^t)\\
    &= \mathrm{Upsample}\left( \mathrm{ResNet}\left( \mathrm{Conv}(\mathbf{u}_{\text{BEV}}^t) \right) \right).
\end{align}
The decoder consists of a lightweight convolutional layer followed by a pre-trained ResNet backbone, which extracts higher-level contextual features. These features are then upsampled and concatenated with shallow features to retain spatial precision during reconstruction. This structure enables accurate generation of BEV-space semantic maps while preserving the geometric and semantic integrity of the scene. By incorporating this auxiliary task, we provide both qualitative and quantitative insights into the structure and informativeness of the learned features. This also helps verify that the BEV feature encoder aligns with traditional perception objectives, improving transparency and interpretability within the DRL pipeline.

\section{Experiments}
\subsection{Experiment Setting}

\subsubsection{Simulation Environment}

Extensive experiments were conducted using the widely adopted CARLA simulator to evaluate the \texttt{ME$^3$-BEV} framework for autonomous driving. We utilized the panoramic cameras provided by CARLA to generate the necessary panoramic image inputs for the \texttt{ME$^3$-BEV} framework and obtained additional road and traffic information, such as road type, speed limits, traffic light states, and planned waypoints, from the simulator. The \texttt{ME$^3$-BEV} framework outputs vehicle control commands to guide the vehicle’s motion. CARLA offers various maps to simulate urban scenarios with different traffic conditions, the model’s performance was then evaluated across seven maps, Town01 (T1) to Town07 (T7).

\subsubsection{Evaluation Metrics}

Following existing autonomous driving benchmarks and method~\cite{e2e-cla,jia2024bench2drive}, we adopt a set of metrics to comprehensively assess model performance from multiple perspectives. 

The definitions of the evaluation metrics are as follows. The calculation of these metrics follows the benchmark
 of closed-loop end-to-end autonomous driving~\cite{jia2024bench2drive,li2024think}.

\begin{itemize}
\item \textbf{Driving Score:} overall evaluation of driving performance by considering task completion and penalty violations. 

\item \textbf{Collision Rate:} the probability of collisions occurring during the driving process. 

\item \textbf{Timesteps:} the driving time until a task is completed or fails. 

\item \textbf{Similarity:} the average cosine similarity between the vehicle's movement direction and the direction towards the next planned waypoint. 

\item \textbf{Waypoint Distance:} the average distance between the vehicle's current position and the next planned waypoint. 

\item \textbf{Efficiency:}  driving efficiency through the ratio of the vehicle's speed to the surrounding vehicles' speed. 

\item \textbf{Comfortness:} assessing by checking whether the vehicle's longitudinal acceleration, maximum absolute lateral acceleration, yaw rate, yaw acceleration, longitudinal component of jerk, and maximum magnitude of the jerk vector fall within predefined standard intervals, thus indicating the comfort level of driving behavior.

\end{itemize}
\subsubsection{Implementation Details}

Each camera captures images at a resolution of 128×352 pixels, with a temporal input sequence length of 4. The Spatial-Semantic Aggregator transforms the panoramic images into high-dimensional 200×200 BEV representation. We implement the PPO algorithm using Stable-Baselines3~\cite{raffin2021stable} to train our DRL network. A step decay learning rate schedule is adopted, starting with an initial learning rate of $10^{-3}$. The autonomous driving policy is trained in the CARLA Town03 map under low-density traffic conditions. The DRL model was trained using a single NVIDIA RTX 3090 GPU. Each test episode lasts for 128 timesteps and terminates immediately upon a collision. All evaluation metrics are averaged over 100 test episodes. Two traffic density settings are used for evaluation: the low-density scenario includes 50 vehicles and 50 pedestrians, while the high-density scenario consists of 100 vehicles and 100 pedestrians.

\subsection{Convergence Behavior of Training Process}

\begin{figure}
    \centering
    \includegraphics[width=0.8\linewidth]{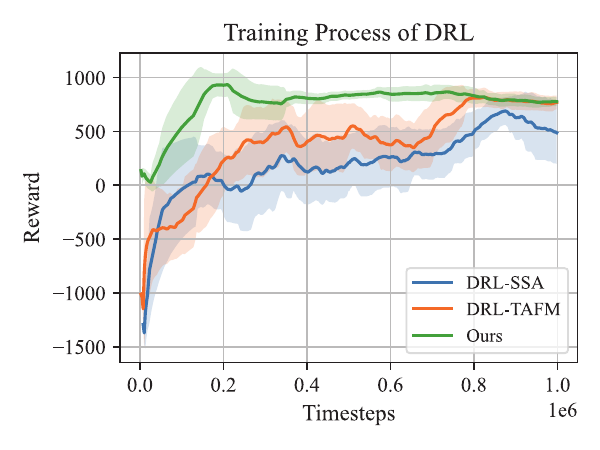}
    \caption{Cumulative rewards change during the training process of DRL as the training environment evolves.}
    \label{fig:reward}
\end{figure}

To evaluate the learning progress and convergence behavior of the proposed deep reinforcement learning-based \texttt{ME$^3$-BEV} framework, we analyze the cumulative reward curves during training and compare them with two variants: DRL-SSA (with only the SSA module) and DRL-TAFM (with only the TAFM module). As shown in Figure~\ref{fig:reward}, the x-axis represents the number of environment steps, while the y-axis denotes the average cumulative reward per step. A moving average is applied to reduce variance and highlight the convergence trend.

The cumulative reward of \texttt{ME$^3$-BEV} increases steadily and stabilizes after approximately 400,000 steps, indicating faster convergence. In contrast, DRL-SSA and DRL-TAFM exhibit slower reward accumulation, converging only after around 800,000 steps and showing noticeable oscillations in the later stages of training. Compared to these baselines, the proposed model achieves not only faster convergence but also significantly reduced reward fluctuations.
These results demonstrate that the proposed architecture facilitates more efficient learning and improves training stability.

\subsection{Performance Evaluation on Seven Carla Maps}
\begin{table*}[]
\small
\centering
\setlength{\tabcolsep}{1mm}
\caption{Comparison of the performance of advanced end-to-end methods based on deep reinforcement learning in terms of autonomous driving control in the low-density and high-density traffic flow of CARLA}
\begin{tabular}{c|c|cccccccc|cccccccc}
\hline
\label{tab:comp}
\multirow{2}{*}{Matric} &
  \multirow{2}{*}{Method} &
  \multicolumn{8}{c|}{Low-density Traffic} &
  \multicolumn{8}{c}{High-density Traffic} \\ \cline{3-18} 
 &
   &
  T1 &
  T2 &
  T3 &
  T4 &
  T5 &
  T6 &
  T7 &
  Avg. &
  T1 &
  T2 &
  T3 &
  T4 &
  T5 &
  T6 &
  T7 &
  Avg. \\ \hline
\multirow{2}{*}{Collision Rate $\left( \downarrow \right)$} &
  \texttt{e2e-CLA} &0.94 &0.92 &0.92 &0.71 &0.83 &0.66 &0.69 &0.81 
  &
  0.95 &
  0.92 &
  0.91 &
  0.82 &
  0.94 &
  0.66 &
  0.79 &
  0.86 \\
 &
  \texttt{ME$^3$-BEV} &\textbf{\underline{0.33}} &\textbf{\underline{0.59}} &\textbf{\underline{0.18}} &\textbf{\underline{0.13}} &\textbf{\underline{0.15}} &\textbf{\underline{0.12}} &\textbf{\underline{0.34}} &\textbf{\underline{0.26}} &\textbf{\underline{0.60}} &\textbf{\underline{0.60}} &\textbf{\underline{0.36}} &\textbf{\underline{0.24}} &\textbf{\underline{0.34}} &\textbf{\underline{0.25}} &\textbf{\underline{0.65}} &\textbf{\underline{0.43}}  \\\hline
\multirow{2}{*}{Similarity $\left( \uparrow \right)$} &
  \texttt{e2e-CLA} &0.78 &0.69 &0.67 &\textbf{\underline{0.66}} &0.64 &\textbf{\underline{0.67}} &0.58 &0.67
  &
  0.73 &
  0.56 &
  0.63 &
  \textbf{\underline{0.68}} &
  0.60 &
  \textbf{\underline{0.62}} &
  0.61 &
  0.63 \\
 &
  \texttt{ME$^3$-BEV} &\textbf{\underline{0.84}} &\textbf{\underline{0.71}} &\textbf{\underline{0.69}} &0.63 &\textbf{\underline{0.83}} &0.65 &\textbf{\underline{0.63}} &\textbf{\underline{0.71}} 
  &
  \textbf{\underline{0.78}} &
  \textbf{\underline{0.64}} &
  \textbf{\underline{0.67}} &
  0.63 &
  \textbf{\underline{0.71}} &
  0.60 &
  \textbf{\underline{0.71}} &
  \textbf{\underline{0.68}} \\ \hline
\multirow{2}{*}{Timesteps $\left( \uparrow \right)$} &
  \texttt{e2e-CLA} &64 &63 &64 &94 &89 &96 &82 &79  
  &
  61 &
  55 &
  64 &
  84 &
  87 &
  96 &
  \textbf{\underline{73}} &
  74 \\
 &
  \texttt{ME$^3$-BEV} &\textbf{\underline{101}} &\textbf{\underline{73}} &\textbf{\underline{113}} &\textbf{\underline{118}} &\textbf{\underline{119}} &\textbf{\underline{115}} &\textbf{\underline{106}} &\textbf{\underline{106}} 
  &\textbf{\underline{68}} &
  \textbf{\underline{63}} &
  \textbf{\underline{87}} &
  \textbf{\underline{105}} &
  \textbf{\underline{100}} &
  \textbf{\underline{99}} &
  60 &
  \textbf{\underline{83}} \\ \hline
\multirow{2}{*}{Waypoint Distance $\left( \downarrow \right)$} &
  \texttt{e2e-CLA} &8.63 &13.24 &7.76 &\textbf{\underline{2.94}} &\textbf{\underline{9.50}} &4.96 &4.66 &7.38
  &
  14.8 &
  32.1 &
  14.2 &
  \textbf{\underline{3.85}} &
  \textbf{\underline{5.44}} &
  \textbf{\underline{4.80}} &
  6.40 &
  11.6 \\
 &
  \texttt{ME$^3$-BEV} &\textbf{\underline{5.66}} &\textbf{\underline{3.51}} &\textbf{\underline{5.82}} &7.37 &14.9 &\textbf{\underline{4.27}} &\textbf{\underline{2.51}} &\textbf{\underline{6.30}}   
  &\textbf{\underline{8.17}} &
  \textbf{\underline{5.57}} &
  \textbf{\underline{12.0}} &
  4.14 &
  8.91 &
  9.40 &
  \textbf{\underline{5.42}} &
  \textbf{\underline{7.66}} \\ \hline
\multirow{2}{*}{Efficiency $\left( \uparrow \right)$} &
  \texttt{e2e-CLA} &\textbf{\underline{1.56}} &\textbf{\underline{4.30}} &\textbf{\underline{1.48}} &0.33 &\textbf{\underline{1.13}} &0.63 &0.95 &\textbf{\underline{1.48}} 
  &
  2.46 &
  4.32 &
  \textbf{\underline{2.89}} &
  0.36 &
  0.87 &
  0.66 &
  1.78 &
  1.91 \\
 &
  \texttt{ME$^3$-BEV} &1.32 &2.65 &1.14 &\textbf{\underline{0.36}} &1.03 &\textbf{\underline{0.67}} &\textbf{\underline{1.38}} &1.22 
  &\textbf{\underline{3.73}} &
  \textbf{\underline{4.91}} &
  2.50 &
  \textbf{\underline{0.45}} &
  \textbf{\underline{1.05}} &
  \textbf{\underline{0.68}} &
  \textbf{\underline{3.11}} &
  \textbf{\underline{2.35}} \\ \hline
\multirow{2}{*}{Comfortness $\left( \uparrow \right)$} &
  \texttt{e2e-CLA} &\textbf{\underline{2.48}} &\textbf{\underline{1.37}} &\textbf{\underline{1.63}} &1.29 &\textbf{\underline{1.48}} &\textbf{\underline{1.32}} &1.65 &\textbf{\underline{1.60}}
  &
  \textbf{\underline{2.48}} &
  \textbf{\underline{1.67}} &
  \textbf{\underline{1.97}} &
  \textbf{\underline{2.07}} &
  \textbf{\underline{1.27}} &
  \textbf{\underline{1.40}} &
  \textbf{\underline{2.27}} &
  \textbf{\underline{1.88}} \\
 &
  \texttt{ME$^3$-BEV} &1.12 &1.06 &1.03 &\textbf{\underline{1.46}} &1.05 &1.10 &1.10 &1.13   
  &0.62 &
  0.79 &
  0.63 &
  0.91 &
  0.83 &
  0.87 &
  0.41 &
  0.72 \\ \hline
\multirow{2}{*}{Driving Score $\left( \uparrow \right)$} &
  \texttt{e2e-CLA} &32.6 &31.6 &31.6 &48.0 &44.8 &48.7 &40.6 &39.7
  &30.3 &
  26.3 &
  31.7 &
  42.4 &
  43.0 &
  48.3 &
  35.7 &
  36.8 \\
 &
  \texttt{ME$^3$-BEV} &\textbf{\underline{64.5}} &\textbf{\underline{44.6}} &\textbf{\underline{85.2}} &\textbf{\underline{78.6}} &\textbf{\underline{73.2}} &\textbf{\underline{84.4}} &\textbf{\underline{67.7}} &\textbf{\underline{71.2}}   
  &
  \textbf{\underline{42.1}} &
  \textbf{\underline{36.9}} &
  \textbf{\underline{66.1}} &
  \textbf{\underline{69.3}} &
  \textbf{\underline{62.1}} &
  \textbf{\underline{73.6}} &
  \textbf{\underline{36.2}} &
  \textbf{\underline{55.2}} \\ \hline
\end{tabular}
\end{table*}

We compare \texttt{ME$^3$-BEV} with the state-of-the-art deep reinforcement learning-based end-to-end driving method \texttt{e2e-CLA} across seven CARLA maps. Specifically, \texttt{ME$^3$-BEV} demonstrates significantly lower collision rates, longer timesteps, and higher driving scores across all maps. As shown in Table~\ref{tab:comp}, \texttt{ME$^3$-BEV} consistently outperforms \texttt{e2e-CLA} on most evaluation metrics. It achieves a significantly lower average collision rate of 0.26 compared to 0.81, representing a 68\% reduction and indicating improved safety. The average Timesteps reaches 106 for \texttt{ME$^3$-BEV}, substantially higher than the 79 timesteps of \texttt{e2e-CLA}, suggesting better stability and robustness.

In terms of Driving Score, which reflects overall task performance, \texttt{ME$^3$-BEV} achieves an average of 71.2, outperforming \texttt{e2e-CLA}'s 39.7 by a large margin. For trajectory quality, \texttt{ME$^3$-BEV} yields a higher similarity score of 0.71 and a lower waypoint distance of 6.30, compared to 0.67 and 7.38 for \texttt{e2e-CLA}, indicating more accurate and consistent path following.

While \texttt{ME$^3$-BEV} shows slightly lower efficiency and comfort values (1.22 and 1.13) than \texttt{e2e-CLA} (1.48 and 1.60), this reflects a safer and more conservative driving style, which is preferred in autonomous driving scenarios that prioritize reliability and rule compliance.

To evaluate the generalization and robustness of \texttt{ME$^3$-BEV}, we compare it with the state-of-the-art end-to-end method \texttt{e2e-CLA} under high-density traffic conditions in CARLA, as shown in Table \ref{tab:comp}. \texttt{ME$^3$-BEV} achieves a substantially lower average collision rate (0.43) compared to \texttt{e2e-CLA} (0.86), demonstrating strong robustness despite a moderate increase from its low-density performance (0.26). In particularly challenging environments such as Town04 and Town05, it maintains low collision rates (0.24 and 0.34), indicating effective risk control under congestion.

Although the average episode length decreases under high-density settings, \texttt{ME$^3$-BEV} still achieves 83 timesteps on average, which is notably higher than \texttt{e2e-CLA} (74), reflecting its resilience and stability. The Driving Score also remains superior (55.2 vs. 36.8), despite a drop from the low-density setting (71.2), confirming consistent overall performance in complex scenarios.

In terms of trajectory quality, \texttt{ME$^3$-BEV} outperforms \texttt{e2e-CLA} with a higher Similarity score (0.68 vs. 0.63) and a lower Waypoint Distance (7.66 vs. 11.6), suggesting more precise path following. Notably, the Efficiency improves to 2.35 under high-density traffic, surpassing both its low-density value (1.22) and that of \texttt{e2e-CLA} (1.91), indicating effective navigation strategies under congestion.

While the Comfort score decreases to 0.72 from 1.13, this trade-off is acceptable given the traffic complexity. In contrast, \texttt{e2e-CLA} achieves slightly better comfort at the cost of significantly reduced safety. These results highlight \texttt{ME$^3$-BEV}'s more balanced and robust performance across key evaluation metrics in demanding traffic conditions.

\begin{table*}[]

\centering
\small
\setlength{\tabcolsep}{1mm}
\caption{Effects of the SSA module and the TAFM module.}

\label{tab:my-table3}
\begin{tabular}{c|l|ccccccc|ccccccc}
\hline
\multirow{2}{*}{metrics}                             & \multicolumn{1}{c|}{\multirow{2}{*}{method}}  & \multicolumn{7}{c|}{Low-density Traffic}                     & \multicolumn{7}{c}{High-density Traffic}                                    \\ \cline{3-16} 
                                                     & \multicolumn{1}{c|}{}                         & T1 & T2 & T3 & T4 & T5 & T6 & T7 & T1 & T2 & T3 & T4 & T5 & T6 & T7                \\ \hline
\multicolumn{1}{c|}{\multirow{3}{*}{Collision Rate $\left( \downarrow \right)$}}                               
& \texttt{ME$^3$-BEV}  &\textbf{\underline{0.33}} &\textbf{\underline{0.59}} &\textbf{\underline{0.18}} &\textbf{\underline{0.13}} &\textbf{\underline{0.15}} &\textbf{\underline{0.12}} &\textbf{\underline{0.34}} &\textbf{\underline{0.60}} &\textbf{\underline{0.60}} &\textbf{\underline{0.36}} &\textbf{\underline{0.24}} &\textbf{\underline{0.34}} &\textbf{\underline{0.25}} &\textbf{\underline{0.65}}\\
\multicolumn{1}{c|}{}                                & \texttt{ME$^3$-BEV} w/o SSA  &0.63 &0.83 &0.87 &0.33 &0.37 &0.30 &0.77 &0.84 &0.87 &0.86 &0.52 &0.73 &0.50 &0.77 \\
\multicolumn{1}{c|}{}                                & \texttt{ME$^3$-BEV} w/o TAFM &0.69 &0.75 &0.82 &0.54 &0.40 &0.30 &0.60 &0.74 &0.81 &0.87 &0.54 &0.55 &0.43 &0.68 \\ \hline
\multicolumn{1}{c|}{\multirow{3}{*}{Similarity $\left( \uparrow \right)$}}                                
& \texttt{ME$^3$-BEV} &\textbf{\underline{0.84}} &\textbf{\underline{0.71}} &\textbf{\underline{0.69}} &\textbf{\underline{0.63}} &\textbf{\underline{0.83}} &\textbf{\underline{0.65}} &0.63 &\textbf{\underline{0.78}} &0.64 &\textbf{\underline{0.67}} &\textbf{\underline{0.63}} &\textbf{\underline{0.71}} &\textbf{\underline{0.60}} &\textbf{\underline{0.71}}  \\
\multicolumn{1}{c|}{}                                & \texttt{ME$^3$-BEV} w/o SSA  &0.73 &0.54 &0.48 &0.53 &0.62 &0.25 &\textbf{\underline{0.79}} &0.59 &0.65 &0.66 &0.59 &0.52 &0.51 &0.56 \\
\multicolumn{1}{c|}{}                                & \texttt{ME$^3$-BEV} w/o TAFM &0.79 &0.57 &0.59 &0.58 &0.53 &0.56 &0.46 &0.72 &\textbf{\underline{0.66}} &0.55 &0.59 &0.62 &0.40 &0.57 \\ \hline
\multicolumn{1}{c|}{\multirow{3}{*}{Timesteps $\left( \uparrow \right)$}}                              
& \texttt{ME$^3$-BEV} &\textbf{\underline{101}} &\textbf{\underline{73}} &\textbf{\underline{113}} &\textbf{\underline{118}} &\textbf{\underline{119}} &\textbf{\underline{115}} &\textbf{\underline{106}} &68 &\textbf{\underline{63}} &\textbf{\underline{87}} &\textbf{\underline{105}} &\textbf{\underline{100}} &\textbf{\underline{99}} &60  \\
\multicolumn{1}{c|}{}                                & \texttt{ME$^3$-BEV} w/o SSA  &74 &64 &54 &114 &114 &107 &77 &59 &34 &25 &102 &81 &90 &59 \\
\multicolumn{1}{c|}{}                                & \texttt{ME$^3$-BEV} w/o TAFM &84 &72 &52 &103 &104 &111 &87 &\textbf{\underline{69}} &43 &43 &88 &90 &98 &\textbf{\underline{64}}  \\ \hline
\multicolumn{1}{c|}{\multirow{3}{*}{Waypoint Distance $\left( \downarrow \right)$}}                              
& \texttt{ME$^3$-BEV} &\textbf{\underline{5.66}} &\textbf{\underline{3.51}} &\textbf{\underline{5.82}} &7.37 &\textbf{\underline{14.9}} &\textbf{\underline{4.27}} &\textbf{\underline{2.51}} &\textbf{\underline{8.17}} &5.57 &\textbf{\underline{12.0}} &\textbf{\underline{4.14}} &8.91 &9.40 &5.42  \\
\multicolumn{1}{c|}{}                                & \texttt{ME$^3$-BEV} w/o SSA  &6.15 &22.6 &12.8 &\textbf{\underline{4.66}} &15.7 &17.8 &14.6 &22.1 &33.9 &17.6 &6.89 &3.59 &8.19 &\textbf{\underline{4.83}} \\
\multicolumn{1}{c|}{}                                & \texttt{ME$^3$-BEV} w/o TAFM &7.96 &3.76 &14.3 &7.81 &21.3 &9.84 &31.4 &6.68 &\textbf{\underline{3.81}} &25.4 &12.3 &19.8 &\textbf{\underline{5.84}} &5.77  \\ \hline
\multicolumn{1}{c|}{\multirow{3}{*}{Efficiency $\left( \uparrow \right)$}}                               
& \texttt{ME$^3$-BEV} &1.32 &\textbf{\underline{2.65}}&1.14 &0.36 &1.03 &0.67 &1.38 &3.73 &\textbf{\underline{4.91}} &2.50 &0.45 &1.05 &0.68 &3.11  \\
\multicolumn{1}{c|}{}                                & \texttt{ME$^3$-BEV} w/o SSA  &1.73 &1.21 &5.18 &0.18 &0.43 &0.83 &\textbf{\underline{2.86}} &\textbf{\underline{3.82}} &2.38 &2.19 &0.69 &\textbf{\underline{3.36}} &1.51 &\textbf{\underline{5.45}} \\
\multicolumn{1}{c|}{}                                & \texttt{ME$^3$-BEV} w/o TAFM &\textbf{\underline{2.26}} &2.31 &\textbf{\underline{5.23}} &\textbf{\underline{0.61}} &\textbf{\underline{1.80}} &\textbf{\underline{0.91}} &1.96 &2.28 &4.32 &\textbf{\underline{4.65}} &\textbf{\underline{0.81}} &1.96 &\textbf{\underline{1.59}} &5.23  \\ \hline
\multicolumn{1}{c|}{\multirow{3}{*}{Comfortness $\left( \uparrow \right)$}}                                
& \texttt{ME$^3$-BEV} & 1.12&1.06 &1.03 &1.46 &1.05 &1.10 &1.10 &0.62 &\textbf{\underline{0.79}} &\textbf{\underline{0.63}} &\textbf{\underline{0.91}} &0.83 &\textbf{\underline{0.87}} &0.41  \\
\multicolumn{1}{c|}{}                                & \texttt{ME$^3$-BEV} w/o SSA  &\textbf{\underline{1.78}} &\textbf{\underline{1.98}} &\textbf{\underline{1.20}} &\textbf{\underline{1.91}} &\textbf{\underline{2.01}} &\textbf{\underline{1.52}} &\textbf{\underline{2.18}} &0.36 &0.17 &0.17 &0.69 &\textbf{\underline{0.91}} &0.77 &\textbf{\underline{0.65}} \\
\multicolumn{1}{c|}{}                                & \texttt{ME$^3$-BEV} w/o TAFM &1.11 &1.20 &1.11 &1.25 &1.04 &0.97 &1.24 &\textbf{\underline{0.95}} &0.41 &0.61 &0.80 &0.81 &0.72 &0.60  \\ \hline
\multicolumn{1}{c|}{\multirow{3}{*}{Driving Score $\left( \uparrow \right)$}}                               
& \texttt{ME$^3$-BEV} &\textbf{\underline{64.5}} &\textbf{\underline{44.6}} &\textbf{\underline{85.2}} &\textbf{\underline{78.6}} &\textbf{\underline{73.2}} &\textbf{\underline{84.4}} &\textbf{\underline{67.7}} &\textbf{\underline{42.1}} &\textbf{\underline{36.9}} &\textbf{\underline{66.1}} &\textbf{\underline{69.3}} &\textbf{\underline{62.1}} &\textbf{\underline{73.6}} &\textbf{\underline{36.2}}  \\
\multicolumn{1}{c|}{}                                & \texttt{ME$^3$-BEV} w/o SSA  &44.0 &32.2 &19.2 &69.8 &68.3 &71.0 &42.2 &32.2 &18.3 &13.1 &61.9 &40.1 &58.5 &32.9 \\
\multicolumn{1}{c|}{}                                & \texttt{ME$^3$-BEV} w/o TAFM &36.7 &29.9 &25.0 &46.5 &44.1 &47.9 &37.6 &29.8 &18.3 &20.7 &39.2 &37.0 &42.4 &25.6  \\

\hline
\end{tabular}%
\end{table*}

We conduct ablation studies on the two core components of \texttt{ME$^3$-BEV}, SSA and TAFM, by removing them individually to assess their contributions to autonomous driving performance. Results are presented in Table~\ref{tab:my-table3}.

SSA has a strong impact on collision rate, driving score, and comfort, especially under high-density traffic. For example, in Town03, the collision rate increases from 0.36 to 0.86, and in Town05, from 0.34 to 0.73. The driving score drops significantly, from 66.1 to 13.1 in Town03. These findings suggest that SSA improves spatial perception by transforming panoramic inputs into BEV features, enabling the policy to better recognize surrounding obstacles and navigate safely. Comfort also declines without SSA, indicating its role in supporting smoother and more stable driving.

TAFM primarily affects trajectory accuracy and decision consistency. Without TAFM, the waypoint distance in Town07 increases from 2.51 to 31.4, showing greater deviation from planned paths. Driving scores also decrease under high-density conditions, for example from 66.1 to 20.7 in Town03 and from 62.1 to 37.0 in Town05. These results demonstrate the importance of long-term temporal modeling for reliable and accurate trajectory planning.
Overall, SSA contributes to safer and smoother driving through improved spatial understanding, while TAFM supports accurate trajectory execution by modeling temporal dependencies. Both components are essential for robust driving performance.

\subsection{BEV Feature Visualization}

\begin{figure}[t]
    \centering
    \includegraphics[width=1\columnwidth]{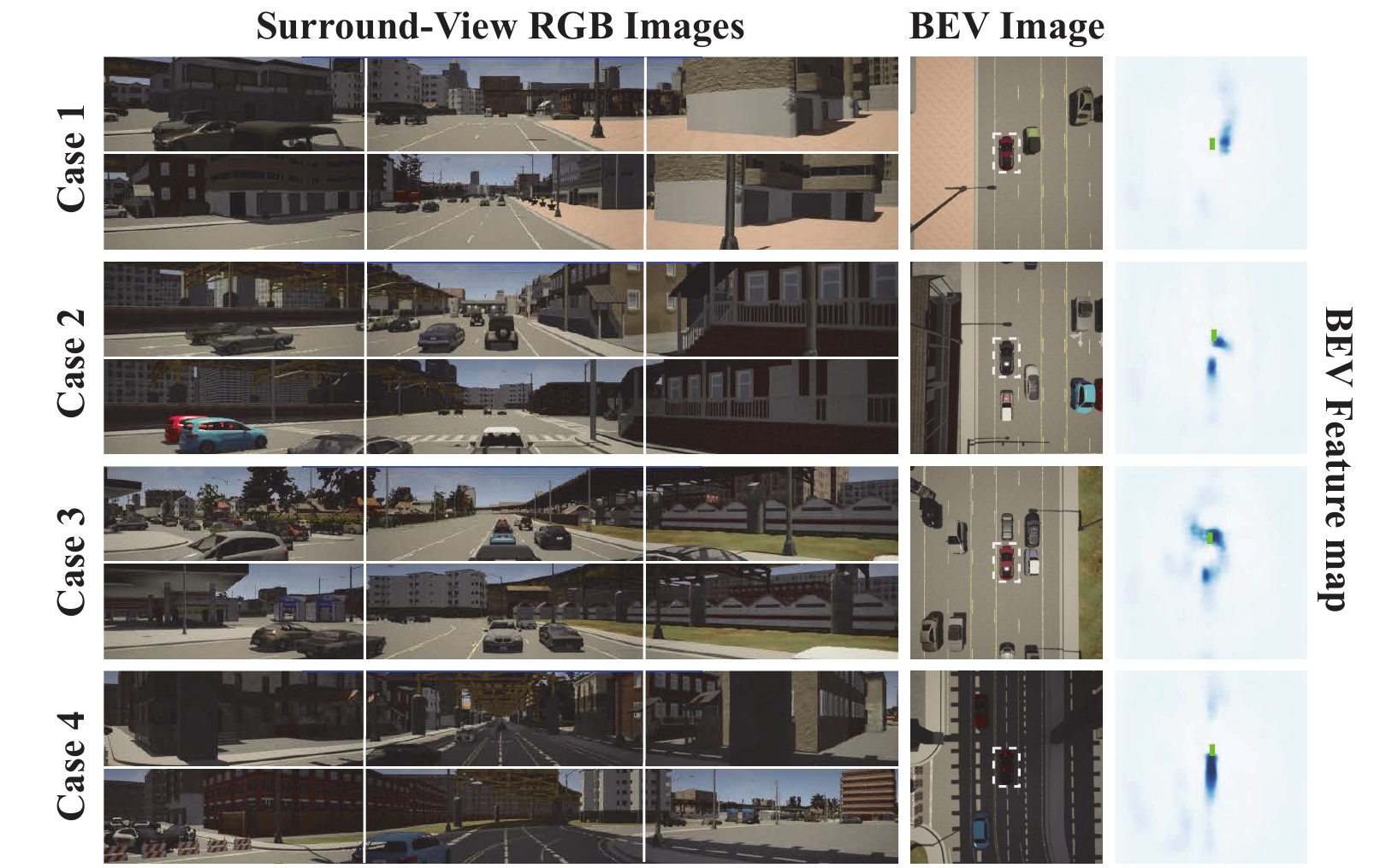}
    \caption{Surround-view images, top-down BEV images and BEV feature maps.}
    \label{fig:explan}
\end{figure}

To evaluate the interpretability of \texttt{ME$^3$-BEV}, we designed a visualization experiment. Specifically, six RGB images captured by surround-view cameras were fed into the model’s feature extraction (perception) network, and the resulting BEV (bird’s-eye view) feature map was generated through a decoder. This BEV representation was then compared with the ground-truth BEV image captured by a top-down camera in CARLA. As shown in Figure~\ref{fig:explan}, the model demonstrates strong multi-view fusion capabilities, effectively integrating information from multiple perspectives into a structurally coherent top-down view.

The BEV feature map produced by the perception network aligns well with the real BEV image in terms of spatial structure, capturing key semantic elements such as road layout, lane markings, and static obstacles. This indicates that the model possesses strong spatial modeling and scene understanding capabilities. The BEV features accurately reflect the spatial distribution of objects and obstacles, thereby providing a reliable and interpretable feature representation to support end-to-end policy learning in autonomous driving.

\section{Conclusion}
In this paper, we presented \texttt{ME$^3$-BEV}, an end-to-end autonomous driving control framework based on deep reinforcement learning. The model integrates an SSA module for enhanced spatial perception and a TAFM module for effective long-term temporal modeling. The SSA module improves the policy’s ability to perceive surrounding traffic participants and avoid collisions, thereby enhancing driving safety. The TAFM module enables efficient extraction of long-horizon temporal features while maintaining fast inference speed, supporting the generation of accurate and consistent driving trajectories and making it suitable for real-time decision-making.
Extensive experiments across diverse CARLA scenarios, including both low- and high-density traffic, show that \texttt{ME$^3$-BEV} consistently outperforms state-of-the-art baselines in safety, trajectory quality, and task completion. Ablation studies further confirm the individual contributions of the SSA and TAFM modules to the model’s robust and interpretable performance.

Looking forward, we plan to extend \texttt{ME$^3$-BEV} to more realistic and dynamic settings. This includes evaluating its generalization capability in real-world urban environments with varying weather, lighting, and traffic conditions, as well as incorporating sensor noise and hardware constraints. 

\section{Acknowledgments}
This work was supported by the National Key R\&D Program of China, 
Project "Development of Large Model Technology and Scenario Library Construction for Autonomous Driving Data Closed-Loop" (Grant No. 2024YFB2505501); the Independent Research Project of the State Key Laboratory of Intelligent Green Vehicle and Mobility, Tsinghua University (no. ZZ-GG-20250405);
and the Guangxi Key Science and Technology Project, Project "Research and Industrialization of High-Performance and Cost-Effective Urban Pilot Driving Technologies" (Grant No.Guangxi Science and Technology AA24206054).
\bibliography{aaai2026}

\end{document}